\documentclass[a4paper,final,10pt,twocolumn]{elsarticle}

\usepackage[top=1.5cm,bottom=2cm,left=1.3cm,right=1.3cm]{geometry}

\usepackage[T1]{fontenc}
\usepackage[utf8]{inputenc}
\usepackage{diagbox}
\usepackage{siunitx}
\usepackage{subcaption}

\usepackage{multirow}

\usepackage{algorithmicx}
\usepackage{algpseudocode}
\usepackage{algorithm}
\usepackage{graphicx} 
\usepackage{amsmath}
\usepackage{mathtools}
\usepackage{hyperref}
\usepackage{color}
\usepackage{amssymb} 
\usepackage[normalem]{ulem} 
 \DeclareMathOperator*{\argmin}{argmin}

\journal{Computer Methods and Programs in Biomedicine}
\let\today\relax

\begin{document}
\begin{frontmatter}

\title{3D Reconstruction and Alignment by Consumer RGB-D Sensors and Fiducial Planar Markers for Patient Positioning in Radiation Therapy}

\author{Hamid Sarmadi\fnref{fn2}}
\ead{hamid.sarmadi@imibic.org}

\author{Rafael~Mu\~noz-Salinas\corref{cor1}\fnref{fn1,fn2}}
\ead{rmsalinas@uco.es}

 \author{ M.\'Alvaro Berb\'is\fnref{fn3}}
 \ead{a.berbis@htime.org}
 
 \author{Antonio Luna \fnref{fn4}}
 \ead{aluna70@htime.org}
 
\author{R. Medina-Carnicer\fnref{fn1}\fnref{fn1,fn2}}
\ead{rmedina@uco.es}

\cortext[cor1]{Corresponding author}
\fntext[fn1]{Computing and Numerical Analysis Department, Edificio Einstein. Campus de Rabanales, C\'ordoba University, 14071, C\'ordoba, Spain, Tlfn:(+34)957212255}
 \fntext[fn2]{Instituto Maim\'onides de Investigaci\'on en Biomedicina (IMIBIC). Avenida Men\'endez Pidal s/n, 14004, C\'ordoba, Spain, Tlfn:(+34)957213861}
 \fntext[fn3] {Health Time,  Avda Brillante 106, 14012, Córdoba, Spain}
\fntext[fn4] {Health Time,  Clínica las Nieves, Carmelo Torres 2,23007, Ja\'en, Spain}
\begin{abstract}
\textbf{Background and Objective:} Patient positioning is a crucial step in radiation therapy, for which non-invasive methods have been developed based on surface reconstruction using optical 3D imaging. However, most solutions need expensive specialized hardware and a careful calibration procedure that must be repeated over time.This paper proposes a fast and cheap patient positioning method based on inexpensive consumer level RGB-D sensors.
\\
\\
\textbf{Methods:} The proposed method relies on a 3D reconstruction approach that fuses, in real-time, artificial and natural visual landmarks recorded from a hand-held RGB-D sensor. The video sequence is transformed into a set of keyframes with known poses, that are later refined to obtain a realistic 3D reconstruction of the patient. The use of artificial landmarks allows our method to automatically align the reconstruction to a reference one, without the need of calibrating the system with respect to the linear accelerator coordinate system. 
\\
\\
\textbf{Results:}
The experiments conducted show that our method obtains a median of \SI{1}{\centi\meter} in translational error, and  \SI{1}{\degree} of rotational error  with respect to reference pose. Additionally, the proposed method shows as visual output  overlayed poses (from the reference and the current scene) and an error map that can be used to correct the patient's current pose to match the reference pose.
\\
\\
\textbf{Conclusions:}
A novel approach to obtain 3D body reconstructions for patient positioning without requiring expensive hardware or dedicated graphic cards is proposed. The method can be used to align in real time the patient's current pose to a preview pose, which is a relevant step in radiation therapy. 
\end{abstract}

\end{frontmatter}
\section{Introduction}

The process of radiation therapy has two main phases. The first phase is planning, where some type of computed tomography (CT) is done to obtain a volumetric reconstruction of the patient's body. This helps to find the exact position of the tumor that needs to be treated in the patient's body. Treatment is done in the second phase which typically happens in several sessions. At each session, the patient needs to be positioned in the same pose in which the reference CT scan has been done and the exact position of the tumor is calculated by aligning the CT scan with patient's body. The most common way to perform this is by taking X-ray image(s) and alignment of the obtained 2D information with the reference 3D model. This normally needs to be done manually by a specialist. There are other options such as applying cone beam CT (CBCT) to get a 3D model in the treatment session and use that instead of the 2D imaging information. A drawback of these approaches is exposing the healthy tissue of the patient to ionizing radiations (e.g. X-ray) at each session of the therapy.

There are non-invasive methods for patient positioning that do not need healthy tissue to be exposed to ionizing radiation. An important subset of these methods are the ones based on optical imaging that make use of visible light and/or infrared sensors. One solution of this type is the use of infrared cameras and reflective markers \cite{desplanques_comparative_2013}. This approach is similar to motion capture systems that are employed in the movie and gaming industry. The measurements of this type of systems are accurate however they are limited to the points on the patient's body where the reflective infrared markers are attached to. Also, in the case of attaching the markers on a cast \cite{desplanques_comparative_2013}, only the rigid pose of the cast is estimated. There are also solutions that take advantage of stereo reconstruction to make a 3D model of the patient's body surface. The reconstructed body surface could then be aligned with the surface of the body in the reference CT scan to align the patient in the right position. This could be easily done with surface alignment algorithms such as iterative closest point (ICP) without the need of known point-wise correspondences between the two surfaces.

Currently, there are several commercial optical-based patient positioning systems that make use of 3D surface reconstruction such as AlignRT, Catalyst and IDENTIFY \cite{hoisak_role_2018}. These are the so-called Surface Guided Radiation Therapy (SGRT) systems. These methods might not be as good as radiation based positioning methods in all cases \cite{wiencierz_clinical_2016}, however, they are a very good alternative to reduce the number of times that radiation-based methods are done \cite{stieler_clinical_2012}. A drawback of the current commercial SGRT solutions is their high price and service costs which might not be affordable for hospitals with limited budgets. Another disadvantage of these systems is that they require their sensors to be fixed in the environment. Therefore, they need to be periodically calibrated with respect to the environment by specialized staff to assure that they report accurate measurements. 

In recent years, starting with the introduction of Microsoft Kinect, inexpensive RGB-D sensors have become available for normal consumers. Starting with the KinectFusion \cite{newcombe_kinectfusion:_2011} many 3D reconstruction algorithms were introduced employing this type of affordable sensors. Nevertheless, there are very few works using these types of sensor for patient positioning \cite{xia_real-time_2012,tahavori_assessment_2013,ogunmolu_real-time_2015,guillet_use_2014} and their potential is not properly explored despite the fact that these consumer sensors are not as accurate as the ones used in commercial patient positioning systems.

This paper proposes a novel patient positioning method based on affordable consumer handheld RGB-D cameras that employs a Simultaneous Localization and Mapping (SLAM) approach that fuses natural features of the patient's body with a set of artificial fiducial planar markers in order to speed up the reconstruction and positioning process.

The proposed method can be run in a regular computer without special hardware or graphics card and create a complete reconstruction and visualization within and average of 31 seconds based on our experiment. Our experimental results show that the proposed method allows a median accuracy of \SI{1}{\centi\meter}  in translational error and  \SI{1}{\degree} of rotational error for rigid transformation.

The rest of this article is organized as follows. First, we explore the optical-based solutions to patient positioning in radiation therapy in Section \ref{secc::relwork}. Then, in Section \ref{sec::method}, we introduce our method. In Section \ref{sec::exp_discuss}, we present the results obtained by our algorithm and discuss the results. Finally, we present our conclusions in the last section.

\section{Background and Objective}
\label{secc::relwork}
One of the first optical patient positioning systems was \cite{bova_university_1997}, where infrared light emitting diodes were attached on a bite plate and the head pose of the patient was inferred from the 3D position of the diodes employing infrared camera images. Another similar early example is \cite{lappe_computer-controlled_1997} that also uses infrared diodes and cameras. In this case, the system automatically corrects the position of the head by a motorized mechanism that corrects its position by translating it so that the isocenter is focused on the correct position. Later, Ploeger et. al. \cite{ploeger_application_2003} used image matching between a video recorded in the treatment phase and reference CT scan using body contours. They concluded that the outline of the patient's body is a more accurate reference that the markers put on their abdomen.

One of the first works investigating the use of 3D surface imaging is by Bert et. al. \cite{bert_phantom_2005}. They analyze the accuracy of the commercially developed patient positioning system AlignRT, which reconstructs the surface by projecting a speckle pattern on the patient and using active stereo. It needs to be calibrated by a special pattern to the coordinate system of the linear accelerator. The system proves to be of high accuracy in estimating rigid transformation. The tests were done on a human phantom.

Around the same time, Bradly et. al. \cite{bradley_vision_2005} use a stationary multi-line laser projector for 3D reconstruction. They employ the iterative closest point (ICP) algorithm to align the 3D surface obtained from a CT scan to the surface from the optical 3D reconstruction. They employed the cast of a human for evaluation. They concluded that their system is good enough to be used for patient positioning. Bert et. al. \cite{bert_clinical_2006} compared the quality of calculating the displacement by laser alignment, portal imaging and the AlignRT surface imaging system for breast treatment. They found that the 3D surface imaging system has superior results in comparison to portal imaging and laser alignment.

Stieler et. al. \cite{stieler_clinical_2012} evaluated a commercial laser scanner Sentinel (by C-Rad AB, Sweden). They found that it is a good solution for the situations where cone beam CT scan or ultrasound imaging are not used, to improve the accuracy of patient positioning.

Desplanques et. al. \cite{desplanques_comparative_2013} introduced a patient positioning system using a pelvic cast or a face mask with reflective markers attached to them. The cast (or mask) was tracked using an infrared motion tracking system and the positioning corrections were compared to those of a patient verification system employing x-ray radiation. They conclude that due to large uncertainty because of the relative motion between the immobilization devices and the patient the system cannot be used as a primary assessment for the quality of patient positioning.

Gaisberger et. al. \cite{gaisberger_three-dimensional_2013} proposed a non-commercial surface scanning system for patient positioning using two optical projectors and two cameras. They concluded that their 3D surface scanning system is good enough to be a viable alternative to the normal kV image-guided radiation therapy.

Wiencierz et. at. \cite{wiencierz_clinical_2016} compared the performance of AlignRT to another commercially available system named Catalyst (from C-Rad, Sweden). The Catalyst system, similar to  AlignRT, has a depth sensor attached to the ceiling of the radiation room. However, unlike AlignRT, it takes advantage of structured lighting of a stripe pattern instead of projecting a speckle pattern. Furthermore, the Catalyst system has one unit instead of two units. The authors report a better accuracy for AlignRT than for Catalyst. Additionally, both of these systems were shown to have better accuracy than using conventional skin markers. On top of that, both of these solutions have an accurate enough measurement in at least 75 percent of the time. However, they both fail to meet the safe accuracy taking into account the 90 and 95 percentile of the errors.

With the advent of consumer-grade RGB-D sensors such as Microsoft Kinect and Asus Xition, many authors have  proposed algorithms for three-dimensional reconstruction using this type of sensors \cite{newcombe_kinectfusion:_2011,newcombe_dynamicfusion:_2015,whelan_elasticfusion:_2016,munoz-salinas_flexible_2019}. One disadvantage of these methods is that, in general, they require specialized hardware i.e. powerful graphics cards with high amount of on board memory. On the other hand, few authors have employed these sensors in the field of patient positioning despite the detailed reconstructions that can be obtained with such devices. Bauer et. al. \cite{bauer_multi-modal_2011} suggested a system for coarse initial patient positioning by matching 3D features from the surface data. They employed the original Microsoft Kinect sensor to evaluate their algorithm. They conclude that their method is feasible for coarse initial patient positioning before using a finer scale more accurate positioning approach. An important disadvantage of their approach is that the RGB-D sensor needs to be fixed in the environment and calibrated with respect to it. Furthermore, it forces the sensor to be far from the patient (on the ceiling) therefore the error of the sensor becomes too high to obtain high precision. Additionally, extrinsic calibration of the sensor unit has to be repeated in case of moving it.

The most similar approach to ours that we found is a dissertation of Guillet \cite{guillet_use_2014}, who tested the accuracy and reproducibility the KinectFusion algorithm \cite{newcombe_kinectfusion:_2011} for patient positioning using a couple of rigidly attached sensors. They scanned the same phantom multiple times with the Microsoft Kinect and aligned the reconstruction manually on the coarse level and then refined it using the ICP algorithm. One downside of the KinectFusion algorithm is that it needs GPU accelerated computing. Another disadvantage of \cite{guillet_use_2014} is that its camera pose estimation is prone to drifting.  To fix this problem they proposed to attach the two Kinect sensors on a camera rig and move the radiation therapy couch instead of the sensors to have a better reconstruction. However, fixing the sensors limits the possible movements and amount of details that can be captured from different parts of the patient. It also makes the the process slow.
 
This work proposes a novel approach for patient positioning  that overcomes the above mentioned problem. First, our method does not need calibration, since obtains the reference location from a set of planar markers placed in the environment (that should not be moved from session to session). Second, our approach works as a handheld scanner instead of having the cameras fixed, which reduces the required infrastructure and the scanning time. But also, it allows to position the camera very close to the patients, thus achieving the best accuracy that the sensor can provide. Finally, our method  works in a normal CPU and does not require any special graphic card. A complete reconstruction and visualization using our approach can be done in approximately 23 seconds on a laptop with Intel Core-i7 CPU.

\section{Methods}
\label{sec::method}
\begin{figure*}[t]
    \centering
    \includegraphics[width=\textwidth]{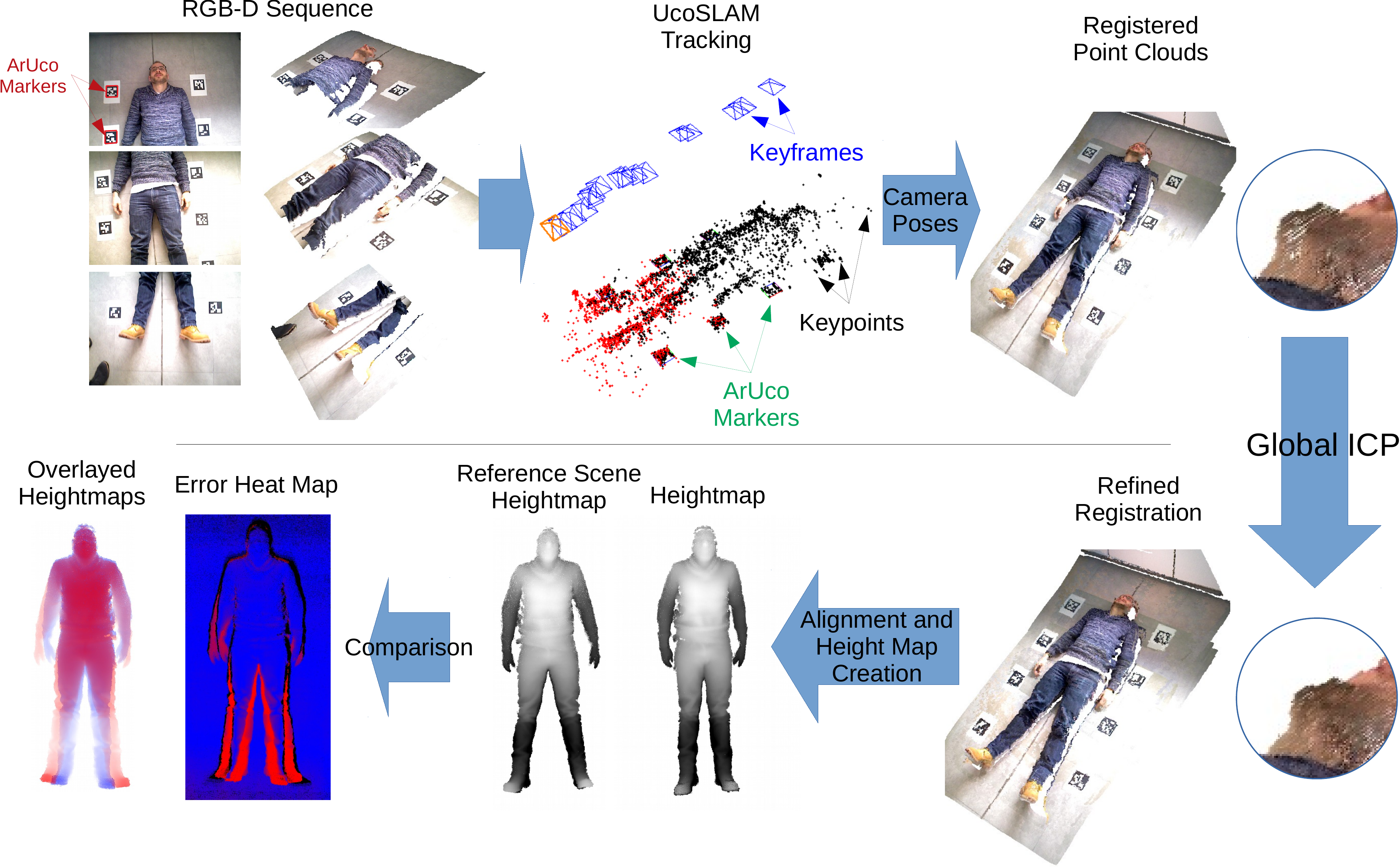}
    \caption{Workflow of the proposed method for patient reconstruction and positioning using a hand-held RGB-D camera. See text for details.}
    \label{fig:example}
\end{figure*}

\begin{figure*}[t]
    \centering
    \includegraphics[width=\textwidth]{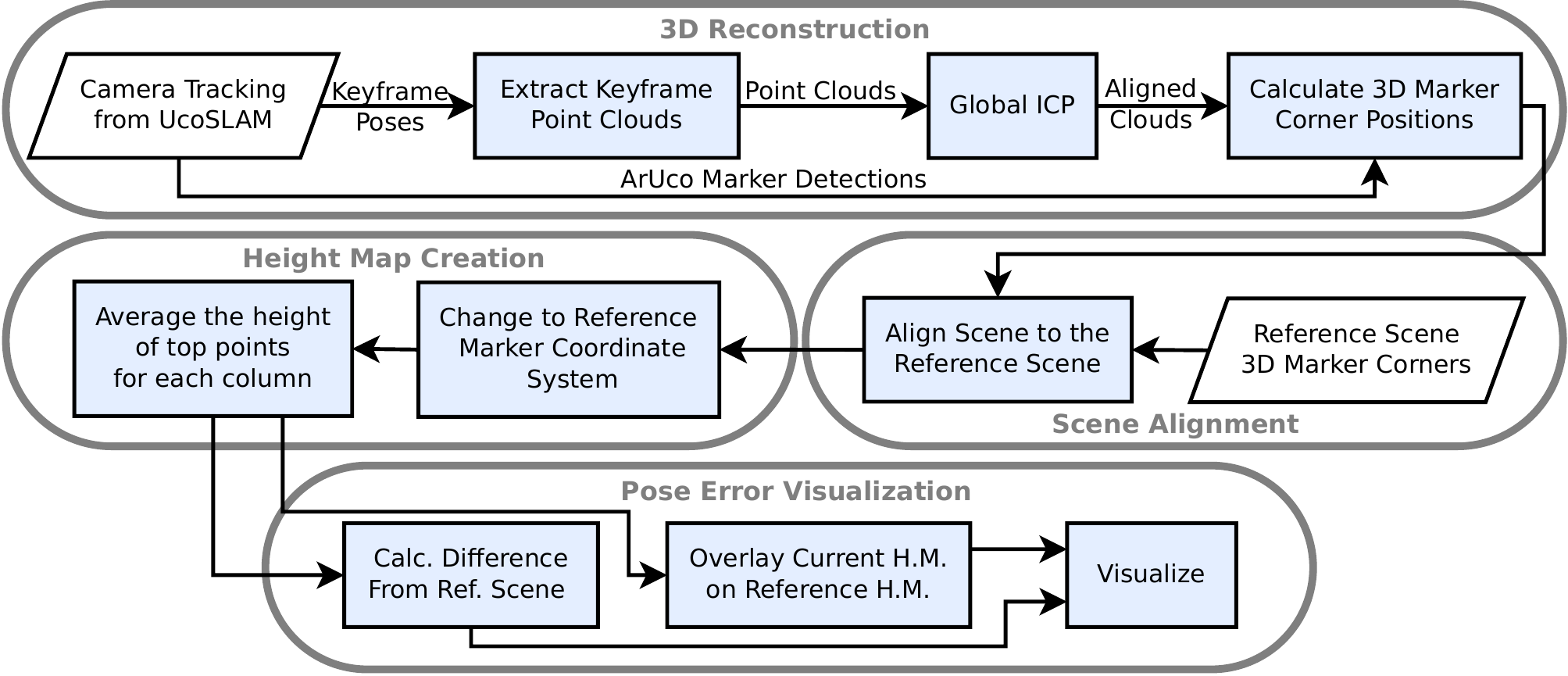}
    \caption{A summary of our proposed algorithm's steps. Please note that Scene Alignment and Pose Error Visualization are only done for non-reference scenes.}
    \label{fig:diagram}
\end{figure*}
\subsection{Overview}
Figure \ref{fig:example} shows a visual summary of our approach. First, an RGB-D sequence is created by scanning the patient using the handheld RGB-D sensor. Then, tracking is done on the RGB-D sequence using the UcoSLAM algorithm \cite{munoz-salinas_ucoslam:_2019} that can take advantage of visual keypoints and also ArUco planar markers \cite{garrido-jurado_generation_2016,ROMERORAMIREZ201838}. The UcoSLAM algorithm generates keyframes and gives camera poses for each keyframe. These are then used to generate registered point clouds for the keyframes. This registration is then refined by a global iterative closest point (ICP) algorithm and the point clouds are converted to a single heightmap. Finally, the heightmap from the current scene is compared to the heightmap from the reference scene to generate an error map and a pose overlay that could be used for the correction of patient's position with respect to the reference scene. Here the current scene is a scene created in the treatment phase of the radiation therapy and the reference scene is the one generated in the planning phase.

Our system takes as input a video sequence recorded (with an RGB-D camera) of the patient from the head to the feet. The video sequence is captured by holding the sensor in hand and moving it over and close to the patient. As already indicated, the environment must have a set of markers placed in arbitrary positions, however they must remain fixed from session to session. The input frames are processed in real-time using a SLAM method able to fuse natural landmarks (keypoints) and the artificial markers. We employ the  UcoSLAM system \cite{munoz-salinas_ucoslam:_2019} developed by the authors of this work. UcoSLAM  simultaneously estimates the camera location and  creates a sparse 3D reconstruction of keypoints and markers. In the process, a set of camera locations are stored, called keyframes, that are later employed in the reconstruction process. After UcoSLAM has processed the recorded sequence, a point cloud is generated for each keyframe using the corresponding depth image from the RGB-D sensor. Then, to refine the relative poses between the point clouds we apply a variant of global iterative closest point (ICP) algorithm \cite{besl_method_1992,zhang_iterative_1992} on all of the clouds. 

One might ask why not using only the ArUco markers to align different point clouds instead of applying the UcoSLAM algorithm. We have two reasons for this. First, there are frames where no markers are visible. By taking advantage of  UcoSLAM, the information from these frames can also be used for reconstruction using purely geometrical information. Second, the color-to-depth registration of the RGB-D camera is not  perfect,  which introduces  errors in the  alignment process. 

The second step of our algorithm is aligning the current reconstructed scene to the reference scene which is created in the planning phase of radiation therapy. For this part,  our method takes advantage of the fiducial markers which we assume that have remained fixed from one recording session to another. 
We would like to remind you that a reference scan needs to be done in the planning phase and markers needs to stay in the same place all through the treatment as in the planning phase. This is done to be able to align the scans in the treatment sessions to the one from the planning session. It is worthy to note that if the CT scan is done in the same room as radiation therapy then it could be aligned to our planning phase surface scan using a registration algorithm such as ICP. In case the CT scan is done in a different room, we still need to make a reference surface scan in the radiation room and align it to the CT scan if we want to know the position of the body internals with respect to the surface reconstruction. One could even choose to do a traditional patient positioning in the planning phase and use the surface scan of that session as the reference. Please note that alignment of the CT scan to the reference scan is not part of our algorithm and could be done by rigid or non-rigid registration algorithms.

After aligning the current scene to the reference scene, a heightmap is obtained from the 3D body reconstruction by averaging the height of the merged point clouds. This process allows to reduce noise, obtaining a smoothed version of the reconstructed surface. 

Finally, when the scenes are aligned and the heightmaps generated, they can be employed to visualize the patient's pose difference between the sessions. Since the heightmaps are aligned together we can subtract them from each other or overlay them on top of each other to create a visualization of the error in positioning the patient.

The rest of the this section provides a formal description of the proposed method.



\subsection{3D Reconstruction}

\noindent Let us define a rigid pose, $P$, as the combination of a rotation matrix, $R$, and a translation vector, $t$, in the 3D space, i.e.:
\begin{equation}
    P=(R,t).
\end{equation}
Now we take:

\begin{equation}
\mathcal\{{P}^{j}_{i}\}, j\in\{1,\cdots,n_i\}
\end{equation}

\noindent as the set of all keyframe poses returned from the UcoSLAM method for the scene $i\in\{0,\cdots,n\}$. A pixel $\mathbf{q}$ in keyframe $j$ of scene $i$  may or may not have a valid depth value $d^{j}_{i}(\mathbf{q})$. Then, let $D^{j}_{i}$ be the set of all pixel positions $\mathbf{q}=[q_x,q_y]^\top \in \mathbb{R}^2$ with a valid depth value $d^{j}_{i}(\mathbf{q})$. Furthermore, let us assume:
\begin{equation}
\mathcal\{{M}^{k}_{ij}\}, k\in\{1,\ldots,m_{ij}\}   
\end{equation}
is the set of markers detected in keyframe  $j$ of scene $i$ and $\mathbf{c}^{kl}_{ij}, l\in \{1,\ldots,4\}$  the $l$-th 2D-corner of the marker $M^{k}_{ij}$.
\subsubsection{Keyframe Point Cloud Creation}
We generate an initial point cloud $C^{j,0}_{i}$ for each keyframe  $j$ of scene $i$ as follows:

\begin{equation}
C^{j,0}_{i}=\{\mathbf{p} \in \mathbb{R}^3|\mathbf{p}=\Psi^{j}_{i}(\mathbf{q},K) , \mathbf{q}\in D^{j}_{i}\},
\end{equation}
\begin{equation}
\Psi^{j}_{i}(\mathbf{q},K)=K^{-1} \begin{bmatrix}\mathbf{q}\\1\end{bmatrix} d^{j}_{i}(\mathbf{q})
\end{equation}

\noindent where $K$ the 3$\times$3 camera matrix and $\Psi^{j}_{i}$ the back projection function for keyframe  $j$ of scene $i$.

In the next step we apply the transformations obtained from the SLAM algorithm to our point clouds $C^{j,0}_{i}$ to obtain new point clouds $C^{j}_{i}$:

\begin{equation}
    C^{j}_{i}=\Theta(C^{j,0}_{i},P^{j}_{i})
\end{equation}
where:
\begin{equation}
    \Theta(C,T)=\{\theta(\mathbf{p},T)| \mathbf{p}\in C\}
\end{equation}
and:
\begin{equation}
    \theta(\mathbf{p},T)=R\,\mathbf{p}+t\quad \text{for} \quad T=(R,t).
\end{equation}
Here $R$ and $t$ are, respectively, the 3D rotation matrix and translation vector related to the transformation $T$. Furthermore, we define the operator $\cdot$ as the combination operator of two transformations in the following way:
\begin{equation}
    T\cdot T'=(RR',t+t')\quad \text{for}\quad T=(R,t), T'=(R',t').
\end{equation}

\subsubsection{Global ICP}
\label{sec::global_icp}
After obtaining the point clouds $C^{j}_{i}$ related to each keyframe $j$ in the scene $i$, we apply our global ICP on all clouds to refine their registration. 

You can find a formal description of our Global ICP procedure in Algorithm \ref{alg1}. As can be seen, in each iteration, for each point of every cloud, we find a corresponding point from a different cloud that has the closest distance to it than any other point in any other cloud. Then we find standard rigid registration for the correspondences found in that iteration step. This registration is denoted by the \textsc{FindTransform}$(.)$ function in Algorithm \ref{alg1} and is obtained by the Horn's algorithm \cite{horn_closed-form_1987} by fixing the scale parameter.

To speed up the process before applying the iterations we randomly subsample the point clouds using a 3D version of the Poisson-disk sampling algorithm. This is denoted by the \textsc{Subsample}$(.)$ function in Algorithm \ref{alg1}. Furthermore, when finding correspondences we only select points that are closer than a certain distance ($r$) because we assume that the initial registration of our point clouds is roughly correct. We use a fixed number of iterations $N_I$ and start with a predetermined maximum value for $r$, $r_\text{max}$, and linearly decrease it to a predetermined minimum value,  $r_\text{min}$, through the iterations. Now, we can write:
\begin{equation}
\begin{multlined}
    (T_i^1,\ldots,T_i^{n_i}) = \\
    \textsc{GlobalICP}(N_I^G,(C_i^1,\ldots,C_i^{n_i}),\sigma^G,r_{min},r_{max}),\\
    i=1,\ldots,n
\end{multlined}
\end{equation}
which indicates that we obtain the transformations $\{T_i^j$, $j=1\ldots n_i\}$, corresponding to keyframe poses $P^{j}_i$, from \textsc{GlobalICP} by giving the input point clouds $\{C_i^j$, $j=1\ldots n_i\}$. Here $N_I^G$ is the input number of iterations and $\sigma^G$ is the input Poisson subsampling radius and $r_{min}$ and $r_{max}$ are the input values for the parameters with the same name in Algorithm \ref{alg1}.

We apply the obtained transformations on their corresponding point clouds $C^{j}_{i}$ to update them to point clouds $\check C^{j}_{i}$ with refined poses:

\begin{equation}
    \check C^{j}_{i}= \Theta(C^{j}_{i},T^{j}_{i}), \quad i=0,\ldots,n,\quad j =1,\ldots,n_i\,.
\end{equation}

\begin{algorithm}[h]
\caption{Global ICP}
\label{alg1}
\begin{algorithmic}
\Procedure{GlobalICP}{$N_I$,$(C_1,..., C_{N_C})$,$\sigma$,$r_\text{min}$,$r_\text{max}$}

\State $r_\text{step} \gets {(r_\text{max}-r_\text{min})\over N_I}$ 
\Comment{$N_I$: number of iterations}

\State $r\gets r_\text{max}$
\For {$j = 1$  \textbf{to}  $N_C$}\Comment{$N_C$: number of point clouds}
\State $T_j\gets (\text{I}_{3\times3},[0,0,0]^\top)$   \Comment{I: the identity matrix}
    \State $\widetilde C_j\gets$\textsc{Subsample}$(C_j,\sigma)$
\EndFor
\newline

\For {$i = 1$  \textbf{to}  $N_I$}\Comment{Obtain the transformations}
    \For{$j = 1$  \textbf{to}  $N_C$}
    
        \State$L_j \gets \varnothing$
        
        \For{$p \in  \widetilde C_j$}
            \State$S\gets\cup_{k=1}^{N_C} \{q \in \widetilde C_k : k \neq j \land ||p-q||_2<r\}$
            
            \If{$S\neq \varnothing$}
                \State$\hat p\gets \argmin_{q\in S}||p-q||_2$
                
                \State$L_j \gets L_j \cup \{(p,\hat p)\}$
            \EndIf
        \EndFor
    \EndFor
    \For{$j = 1$  \textbf{to}  $N_C$}
        \State$\mathcal{T}\gets\textsc{FindTransfrom}(L_j)$
        
        \State$T_j \gets \mathcal{T} \cdot T_j$
        
        \State$\widetilde{C}_j\gets\Theta(\widetilde{C}_j,\mathcal{T})$
        
    \EndFor
\State $r \gets r-r_\text{step}$
\EndFor
\newline
\State
\Return{ $ (T_1 , T_2, ..., T_{N_C})$}
\EndProcedure
\end{algorithmic}
\end{algorithm}
\subsubsection{3D Marker Corner Positions}
\label{sec::corner_positions}
For each corner of each observed marker in a scene we take the average of 3D positions of the detected corners across all keyframes. We find the 3D position of the corners of detected markers from the point clouds using the depth-RGB registration (obviously we assume that this type of correspondence is available for the RGB-D sensor). We take the obtained mean values as the position of the marker corners in each scene. Then, these marker corners can be used to align two reconstructed scenes together.

Let us take $\mathbf{\check c}^{kl}_{ij}$ as the 3D coordinates corresponding the $l$-th 2D-corner $\mathbf{c}^{kl}_{ij}$ of detected marker  $M^{k}_{ij}$.  If the corner has a valid depth value $d^{j}_i(\mathbf{c}^{kl}_{ij})$, we use that to back project the point to get the 3D coordinates. If not, we take the average of the back-projection of points with a valid depth value in the neighbourhood of the corner:
\begin{equation}
\label{eq:markercornerposition}
  \mathbf{ \check c}^{kl}_{ij}=\left\{
    \begin{array}{l l}
    \Psi^{j}_{i}(\mathbf{c}^{kl}_{ij},K) & \mathbf{c}^{kl}_{ij}\in D^{j}_{i}
    \\
    \langle \Psi^{j}_{i}(\mathbf{q},K) \rangle_{\mathbf{q} \in W( \mathbf{c}^{kl}_{ij},D^{j}_{i},s)} & \mathbf{c}^{kl}_{ij} \notin D^{j}_{i}
    \end{array} 
  \right..
\end{equation}
\noindent where $W( \mathbf{c}^{kl}_{ij},D^{j}_{i},s)$ is the set of all points with a valid depth value within an $s\times s$ region centered at the $\mathbf{c}^{kl}_{ij}$ corner and $\langle.\rangle$ is the averaging operator. Please note that we need to retrieve the depth value using a region because the valid values of the depth maps could be sparse at times.

After assigning the 3D coordinates of the marker corners $\mathbf{\check{c}}^{kl}_{ij}$ for each keyframe $j$, we calculate the coordinates of the corners $\mathbf{\check{c}}^{kl}_i$ for the whole scene $i$ by taking their average:
\begin{gather}
   \mathbf{ \check{c}}^{kl}_i=\langle \theta(\mathbf{\check{c}}^{kl}_{ij},T^{j}_{i}) \rangle_{j\in\text{E}_i^{kl}}\\
\label{eq:scenecorners}
\text{E}_i^{kl}=\{j|W(\mathbf{c}_{ji}^{kl},D^{j}_{i},s)\neq \varnothing\}
\end{gather}

Here E$^{kl}_i$ is the set of indices for all keyframes where $\mathbf{\check {c}}^{kl}_{ij}$ has a valid value. Notice that if $W( \mathbf{c}_{ij}^{kl},D^{j}_{i},s) = \varnothing$, it also indicates that $\mathbf{c}^{kl}_{ij}\notin D_i^j$ and that $\mathbf{\check c}^{kl}_{ij}$ does not have a valid value.
\subsection{Scene Alignment}

Let us say that we want to register a current scene (e.g. scene reconstructed in the treatment phase of radiation therapy) to a reference scene (e.g. scene reconstructed in the planning phase of radiation therapy) which we assume has index $0$, using the corner positions. When all valid 3D corner positions of markers are determined for the scene $i$, we find the rigid transformation that transforms each marker corner in the current scene to its corresponding marker corner in the reference scene. We find this transformation using the algorithm by Horn \cite{horn_closed-form_1987}.

 Let $N_{M_i}$ represent the number of different markers in every scene $i$. Because the number of marker corners in each marker is four, then
\begin{equation}
    \begin{gathered}
    L_i=\{(\mathbf{\check{c}}_i^{kl},\mathbf{\check{c}}_0^{kl})|i \neq 0, \; k=1,\ldots,N_{M_i}, \; l=1,\ldots,4\},\\
    i=1,\ldots,n,
    \end{gathered}
\end{equation}
where $L_i$ is the set of point correspondences we use to find the transformation aligning scene $i$ to the reference scene. This transformation, denoted by $\mathfrak{T}_i$ is calculated by:
\begin{equation}
    \mathfrak{T}_i=\textsc{FindTransform}(L_i)
\end{equation}

Finally, the point cloud of each keyframe aligned to the reference scene can be obtained by:
\begin{equation}
    \hat C^{j}_{i}=\Theta(\check C^{j}_{i},\mathfrak{T}_i), \quad j=1,\ldots,n_i, \quad i\neq0
\end{equation}
For the reference scene, however, this step is not necessary, therefore we can write:
\begin{equation}
    \hat C^{j}_{0}=\check C^{j}_{0}, \quad j=1,\ldots,n_0,
\end{equation}
\subsection{Height Map Creation}
\label{sec::height_map}

In order to create a height map we need to have a plane to create the height map grid. Therefore, a marker is chosen as the reference marker and its plane is taken as the grid plane.

For each scene, we take all of the points from all keyframes  and move them to the coordinate system the reference marker. To do so, a transformation is found to take the points from the reference scene's coordinate system to the one of the reference marker in the reference scene.

Assuming the reference marker has index 0 we have: 

\begin{equation}
        \mathfrak{L}=\{(\mathbf{\check{c}}^{0 l}_0,\mathbf{c}_l)|l=1\ldots 4\}
\end{equation}

Since we are using ArUco markers, the position of the marker corners in the coordinate system of the marker are set to:
\begin{equation}
\begin{array}{l l}
\displaystyle\mathbf{c}_1=(-{\mathbf{l}\over2},-{\mathbf{l}\over2},0),
&\displaystyle\mathbf{c}_2=({\mathbf{l}\over2},-{\mathbf{l}\over2},0),\\
&\\
\displaystyle \mathbf{c}_3=({\mathbf{l}\over2},{\mathbf{l}\over2},0),
& \displaystyle \mathbf{c}_4=(-{\mathbf{l}\over2},{\mathbf{l}\over2},0),
\end{array}
\end{equation}
where $\mathbf{l}$ is the length of the side of the square marker.

Now we can calculate the transformation to the reference marker coordinate system and apply it to the point clouds:
\begin{equation}
\mathfrak{C}^{j}_{i}=\Theta(\hat C^{j}_{i}, \textsc{FindTransform}(\mathfrak{L})),
\end{equation}
for $i=0\ldots n$ and $j=1\ldots n_i$.

Subsequently, we create a grid on the marker plane and to get the height value for each pixel in this grid we take the average height of the points that project to that pixel. Since there might be several surfaces on the line that projects to a pixel we take only the points that are within a certain distance from the point with maximum height and keep their average as the height value.

We define the height map grid corresponding to $\mathfrak{C}^{j}_{i}$ as follows:

\begin{equation}
\label{height_map}
H^{j}_{i}=\left \{\textstyle\left (\left\lfloor {\text{x}-x_\mathrm{min}\over \delta} \right\rfloor,\left\lfloor {\text{y}-y_\mathrm{min}\over \delta} \right\rfloor\right) \left |
\begin{array}{l}
(x,y,z) \in  \mathfrak{C}^{j}_{i} \quad\land\\
x_\text{min}\leq x \leq x_\text{max} \quad\land\\
y_\text{min}\leq y \leq y_\text{max}
\end{array}
\right .
\right \} 
\end{equation}
where $\delta$ is the constant step to create the grid, and $x_\text{min}$, $x_\text{max}$, $y_\text{min}$, and $y_\text{max}$ are constant minimum and maximum values for $x$ and $y$, respectively. The boundaries help to trim the height map to our region of interest.
Finally the height value in each cell of the height map related to $H^{j}_{i}$ is defined as:
\begin{equation}
\label{eq:heightmap_h}
    h_{ij}(x,y)= \langle z \rangle _{z\in \mathfrak{b}_{ij}(x,y) \land (Z^\text{max}_{ij}(x,y)-z)<t}
\end{equation}
where $t$ is a threshold to discard the points tha do not belong to the top surface. Also:
\begin{equation}
    \mathfrak{b}_{ij}(\text{x},\text{y})=\left \{z\left |
    \begin{array}{l}
    (x,y,z) \in \mathfrak{C}^{j}_{i} \quad\land\\
    (\lfloor x \rfloor ,\lfloor y \rfloor) \in H^{j}_{i} \quad\land\\
    (\lfloor x \rfloor ,\lfloor y \rfloor) = (\text{x},\text{y})
    \end{array}
    \right .\right\} 
\end{equation}
and
\begin{equation}
    Z^\text{max}_{ij}(x,y)=\max_{z\in\mathfrak{b}_{ij}(x,y)}z.
\end{equation}
Now to get a single height map for each scene we merge the height maps of that scene in this manner:
\begin{equation}
    H_i=\bigcup_{j=1\ldots n_i}H^{j}_{i}
\end{equation}
\begin{equation}
    h_i(x,y)=\langle h_{ij}(x,y)\rangle_{j=1\ldots n_i}
\end{equation}
where $H_i$ is the height map grid related to the scene $i$ and $h_i(x,y)$ is the height value at grid point $(x,y)$ in scene $i$.

Height maps let us merge the keyframe clouds in the scene in a fast manner and just by averaging the height values in the corresponding grid positions. One might argue that some information is lost in the process of converting the point cloud to the height maps. However, we would argue that since the point clouds are generally created by using the depth sensor held above the patient, the information loss is not very high. Furthermore, the averaging operator creates a smooth surface when converting a point cloud to a height map which reduces the noise on the surface.

\section{Results}
\label{sec::exp_discuss}

This section explains the experiments conducted to validate our proposal.  To record RGB-D sequences we used the Asus Xtion Pro Live sensor employing the  OpenNI2 \footnote{structure.io/openni} Linux driver. We choose this sensor because is it cheap, lightweight and does not need an external power supply, it just needs to be connected to the USB port. 

We evaluated our method both qualitatively and quantitatively. The quantitative evaluation (Sect.~\ref{subsec::quantitative}) aims at analyzing the accuracy of the proposed method in estimating the pose of patients with such an inexpensive RGB-D sensor. To do so,  we used a mannequin of the human torso and a commercial motion capture system from OptiTrack\footnote{www.optitrack.com}. 

Our system provides a visual output that can be employed to easily position the patient from one session to another which is based on our 3D reconstruction. Thus, Sect.~\ref{subsec::qualitative} provides a qualitative evaluation of the system outputs and 3D reconstruction provided by our method.

Finally, Sect.~\ref{subsec::computingtimes} provides an analysis of the computing times required by our proposal. As will be explained, our method produces its output within 31 seconds for the human subjects once the video has been recorded, which takes only $10$ seconds.

\subsection{Quantitative evaluation}
\label{subsec::quantitative}

\begin{figure}[t]
    \centering
    \includegraphics[width=0.47\textwidth]{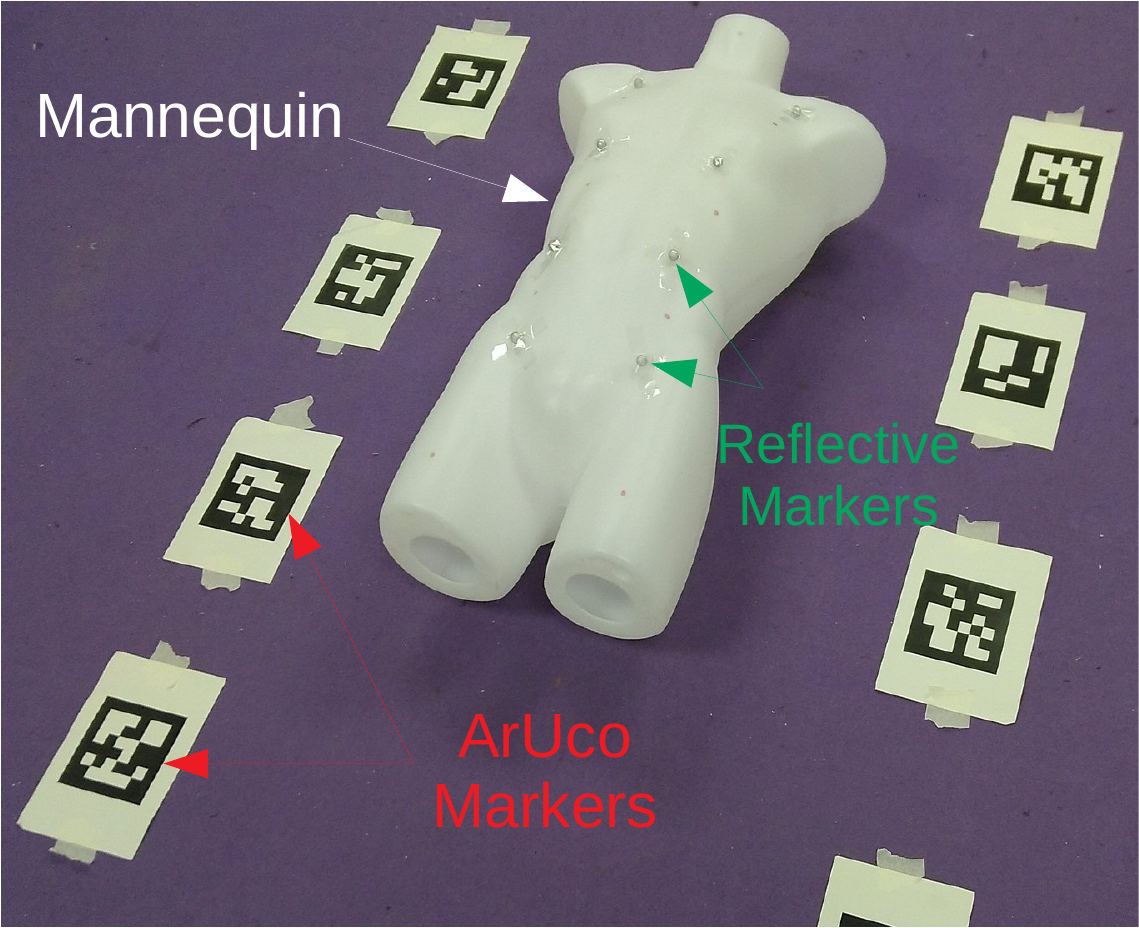}
    \caption{Example of our simulated patient setup.}
    \label{fig:mannquin_setup}
\end{figure}

This section analyzes the precision of our method in estimating the displacement of a 3D body reconstruction with respect to a reference one. For that purpose, a mannequin of the human torso has been placed and scanned at nine different positions on the floor, where multiple ArUco markers were fixed and visible next to the mannequin (see Fig.~\ref{fig:mannquin_setup}).  The parameter values employed for our method are shown in Table \ref{tab:params}.

\begin{table*}[ht!]
\centering
\begin{tabular}{|c|c|p{12cm}|}
\hline
  Parameter& Value &   Description \\
  \hline
  \hline

$N_I^G$ & 20 &  Number of iterations in Algorithm \ref{alg1} (Section \ref{sec::global_icp})\\
\hline

$r_{min}$ & \SI{0.5}{\centi\meter} &  Minimum radius for correspondence association in Algorithm \ref{alg1}. (Section \ref{sec::global_icp})\\
 \hline
 $r_{max}$ & \SI{5}{\centi\meter} &  Maximum radius for correspondence association in Algorithm \ref{alg1}. (Section \ref{sec::global_icp})\\
 \hline
 $\sigma^G$ & \SI{2}{\centi\meter} & Radius for Poisson subsampling the pointcloud.(Section \ref{sec::global_icp})\\
 \hline
 $s$ & \SI{11}{pixels} &  Region length  to calculate depth value where there is not a valid value (Section \ref{sec::corner_positions})\\
\hline 
 $\mathbf{l}$ & \SI{0.104}{\meter} & Side length of each square marker (Section \ref{sec::height_map})\\
\hline 
 $x_{min}$ & \SI{-0.1}{\meter} & Cropping parameter relative to the reference marker to for creating the height map(Section \ref{sec::height_map})\\
 \hline
  $x_{max}$ & \SI{2}{\meter} & Cropping parameter relative to the reference marker to for creating the height map(Section \ref{sec::height_map})\\
 \hline
  $y_{min}$ & \SI{-0.2}{\meter} & Cropping parameter relative to the reference marker to for creating the height map(Section \ref{sec::height_map})\\
 \hline
  $y_{max}$ & \SI{1}{\meter} & Cropping parameter relative to the reference marker to for creating the height map(Section \ref{sec::height_map})\\
 \hline
 $\delta$ & \SI{1.5}{\milli\meter} & Grid step for creating the heightmaps. (Equation \ref{height_map})\\
 \hline
 $t$ & \SI{3}{\centi\meter} & Threshold for picking the top points in a grid cell for creating the heightmaps. (Equation \ref{eq:heightmap_h})\\
 \hline
 
 
  \end{tabular}
\caption{Parameters values used in the quantitative evaluation of our algorithm.}
  \label{tab:params}
 \end{table*} 

The method's precision has been measured as the error in estimating the rigid transformation between two scans of the mannequin. In order to obtain the ground truth, a commercial motion capture system  (OptiTrack) has been employed, which requires to attach several spherical infrared reflective markers on the mannequin surface. The ground truth displacement of the mannequin is calculated by finding the rigid transformation that moves the reflective markers from one scene to another one using Horn's algorithm \cite{horn_closed-form_1987} by fixing the scale parameter.

The same rationale is employed to calculate the rigid transformation with our 3D reconstruction. To do so, the 3D location of the reflective markers is manually extracted from the 3D reconstruction obtained with our method.



Table \ref{tab:results_asus} shows error in estimating the rigid 3D transformation of the mannequin from each scene to every other scene. In other words, each time one scene is taken as the reference scene and the transformation error is calculated with respect to the reference scene from every other scene. Then the mean and median of the rotational and translational errors are calculated for all these other scenes. Also the rotational and translational mean and medians are reported when taking into account all of the data together.

\begin{table}[]
    \centering
    \begin{tabular}{c|r|r|r|r}
    
    \multirow{3}{*}{Reference Scene\#} &
    
    \multicolumn{2}{c|}{\multirow{2}{2cm}{
    Rotation Error (\SI{}{\degree})}} &
    
    \multicolumn{2}{c}{\multirow{2}{2cm}{
    Translation Error (\SI{}{\milli\meter})}}\\
    & \multicolumn{2}{c|}{}& \multicolumn{2}{c}{}\\    \cline{2-5}
    &mean & median& mean& median\\    
    \hline
    0&  1.180 & 0.792 & 8.716 & 8.507\\
    
    1&  1.655 & 1.173 & 11.681 & 9.168\\

    2&  1.305 & 1.073 & 10.704 & 11.169\\

    3&  1.970 & 1.917 & 11.950 & 10.114\\

    4&  1.255 & 0.664 & 10.112 & 10.222\\

    5&  1.330 & 1.005 & 10.783 & 9.063\\

    6&  1.354 & 0.896 & 14.250 & 13.959\\

    7&  3.008 & 3.121 & 11.792 & 10.534\\

    8& 1.348 & 1.103 & 13.135 & 12.610\\

    \hline
    Total & 1.600 & 1.196 & 11.458 & 10.497\\

    \end{tabular}
    \caption{Errors in estimation of the rigid transform for the mannequin from each scene to the reference scene for the dataset captured by the Xtion Pro Live sensor.}
    \label{tab:results_asus}
\end{table}
\subsection{Qualitative evaluation}
\label{subsec::qualitative}
\begin{figure*}[ht!]
    \centering
    \begin{subfigure}[t]{0.25\textwidth}
        \centering
        
        \includegraphics[width=2.1cm]{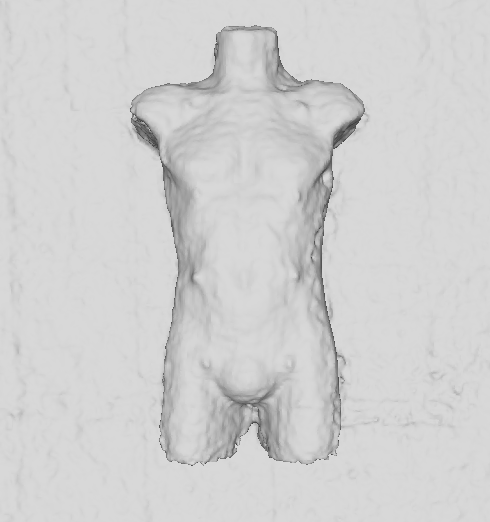}
        \includegraphics[width=2.1cm]{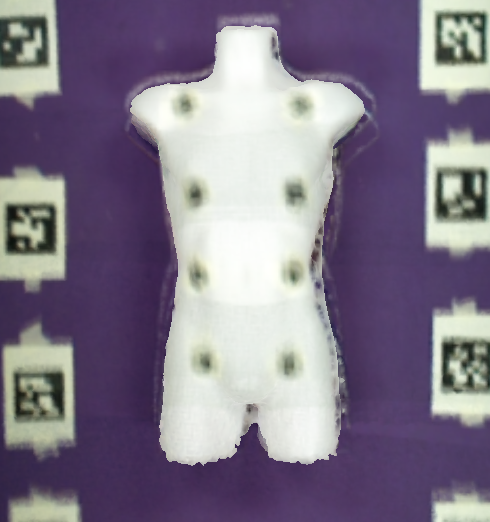}
        
        \includegraphics[width=2.1cm]{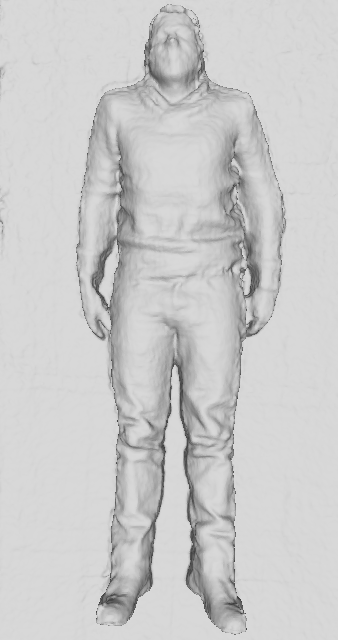}
        \includegraphics[width=2.1cm]{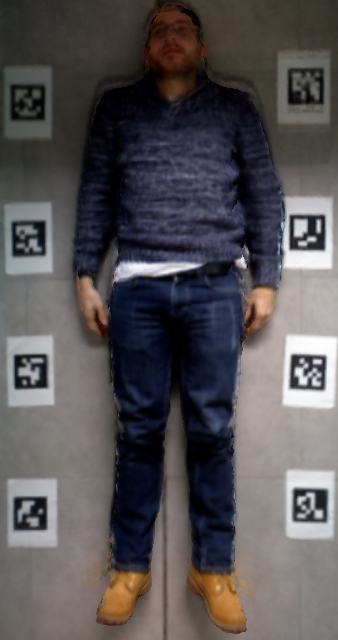}
        
        \includegraphics[width=2.1cm]{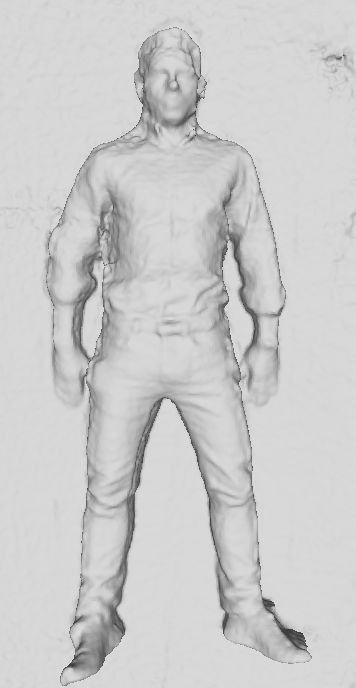}
        \includegraphics[width=2.1cm]{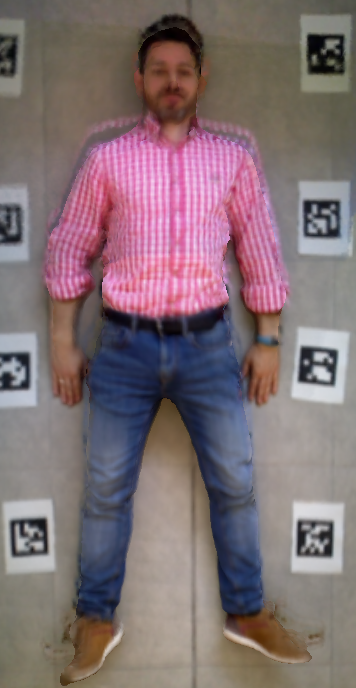}
        
        \includegraphics[width=2.1cm]{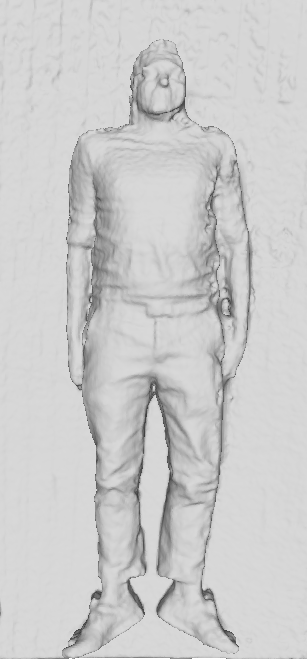}
        \includegraphics[width=2.1cm]{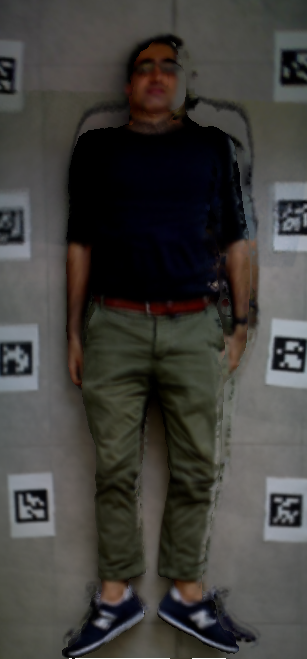}
        \caption{Reference Scene}
    \end{subfigure}
    \begin{subfigure}[t]{0.25\textwidth}
        \centering
        
        \includegraphics[width=2.1cm]{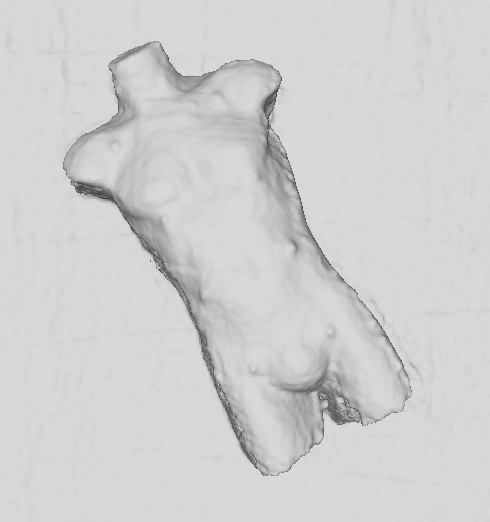}
        \includegraphics[width=2.1cm]{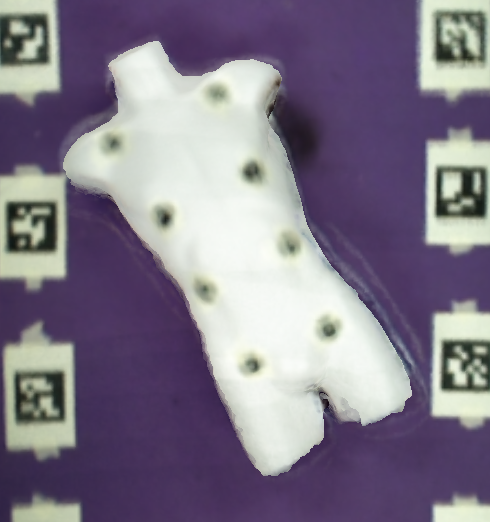}
        
        \includegraphics[width=2.1cm]{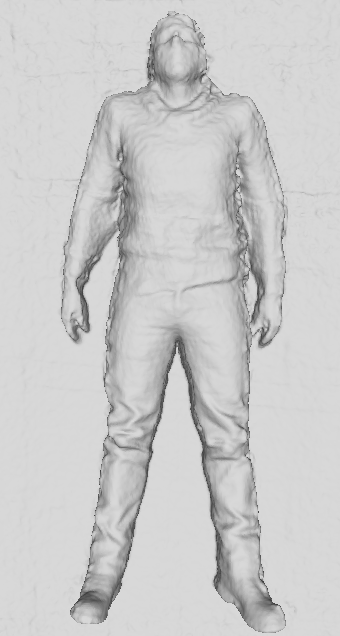}
        \includegraphics[width=2.1cm]{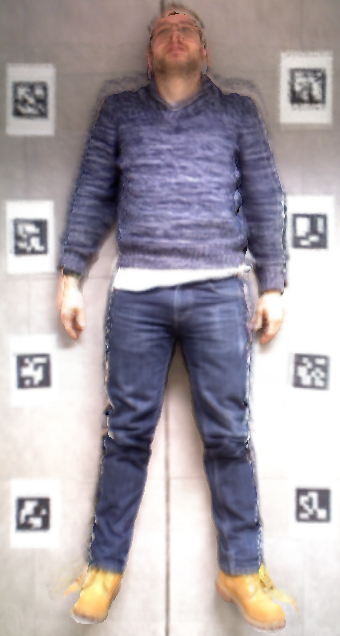}
        
        \includegraphics[width=2.1cm]{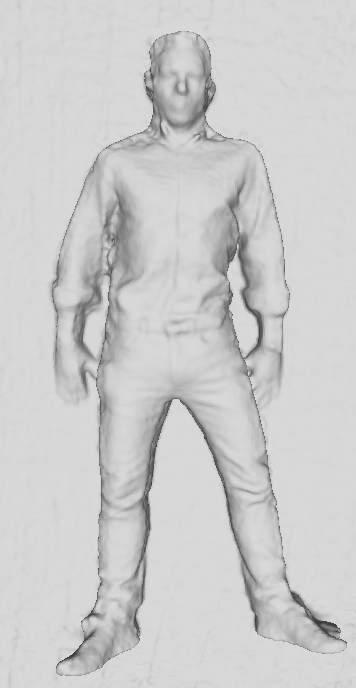}
        \includegraphics[width=2.1cm]{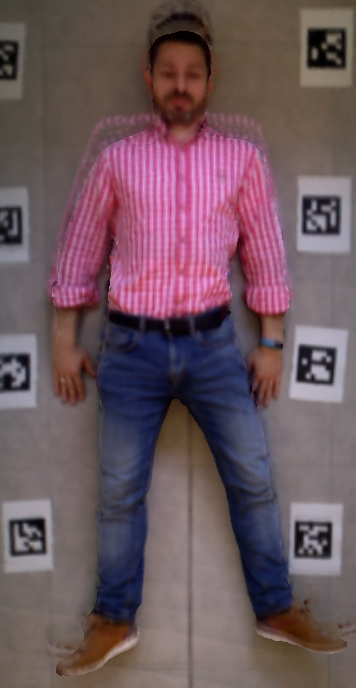}
        
        \includegraphics[width=2.1cm]{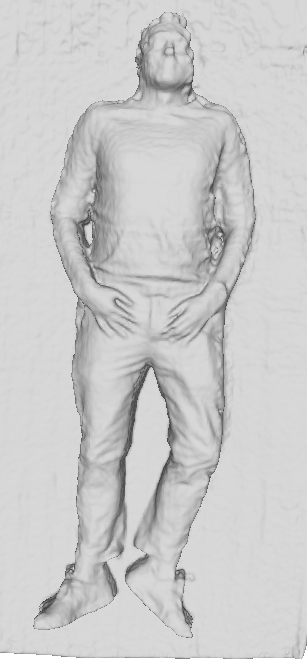}
        \includegraphics[width=2.1cm]{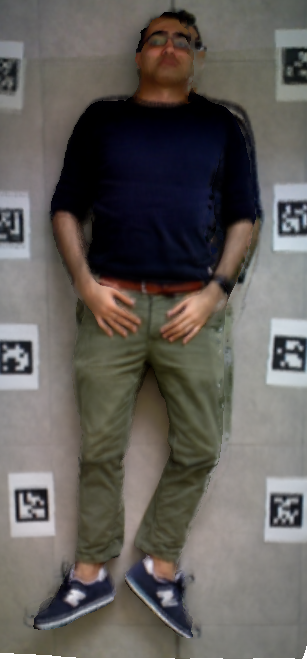}
        \caption{New Scene}
    \end{subfigure}
    \begin{subfigure}[t]{0.24\textwidth}
        \centering
        \includegraphics[width=1.8cm]{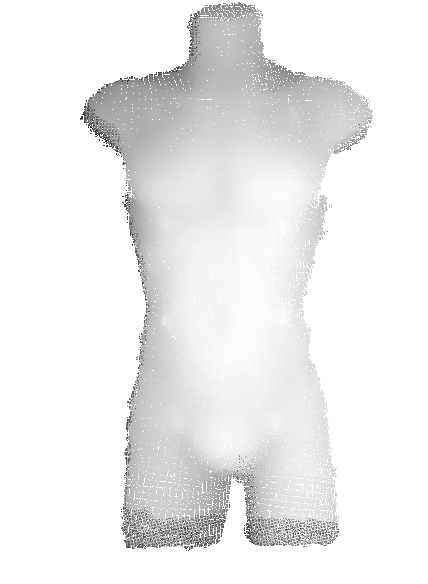}
        \includegraphics[width=1.8cm]{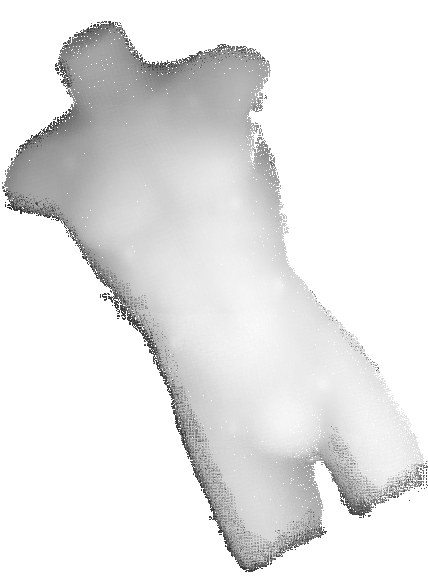}
        
        \includegraphics[width=1.8cm]{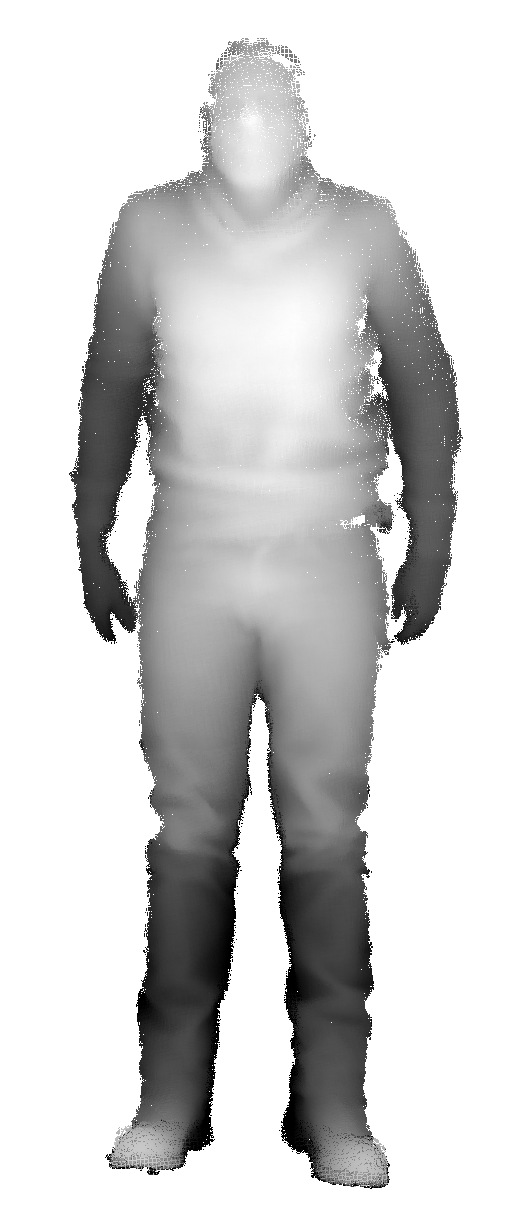}
        \includegraphics[width=1.8cm]{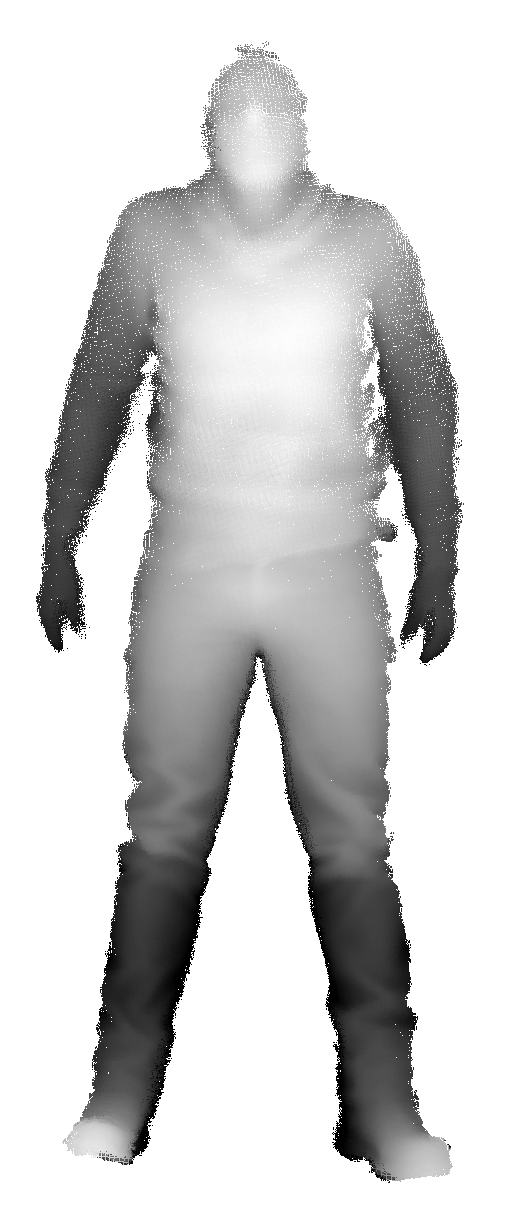}

        \includegraphics[width=1.8cm]{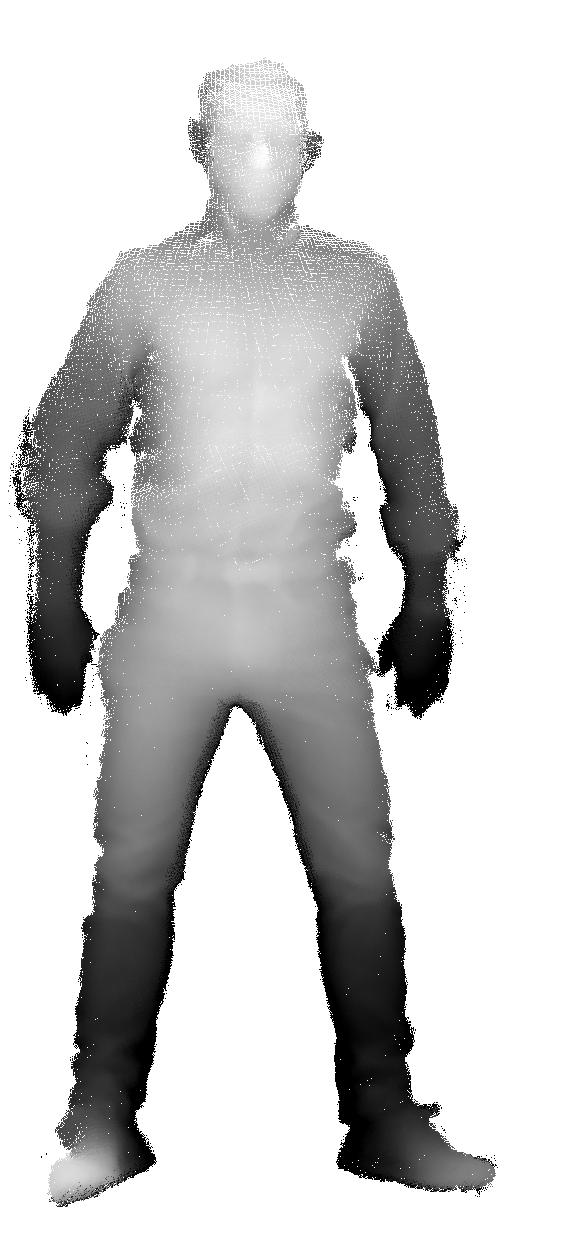}
        \includegraphics[width=1.8cm]{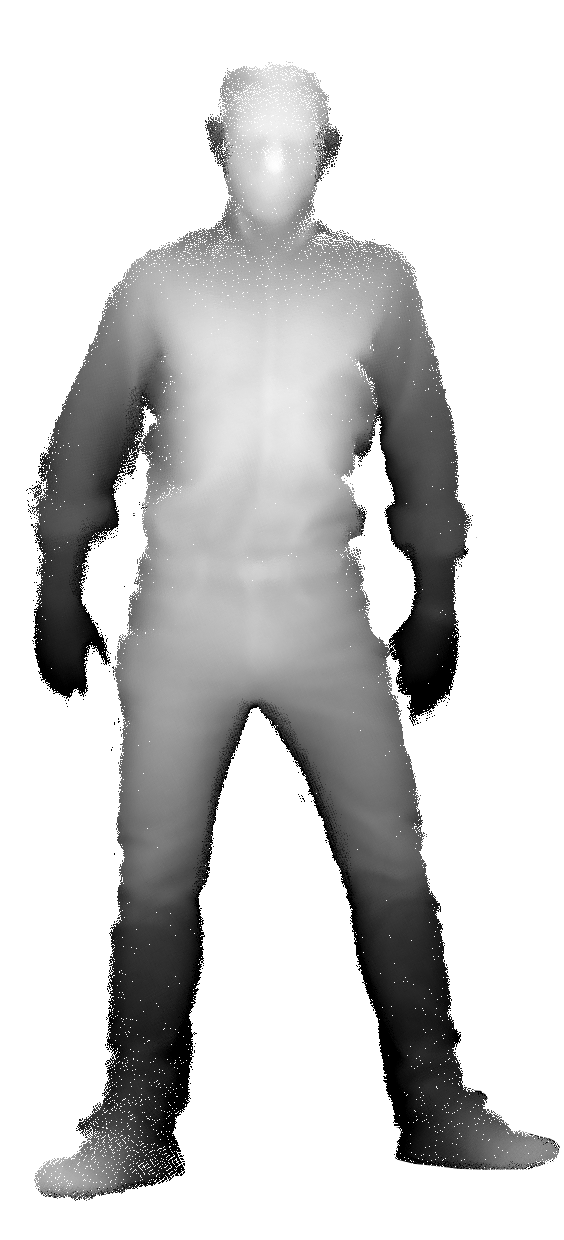}

        \includegraphics[width=1.8cm]{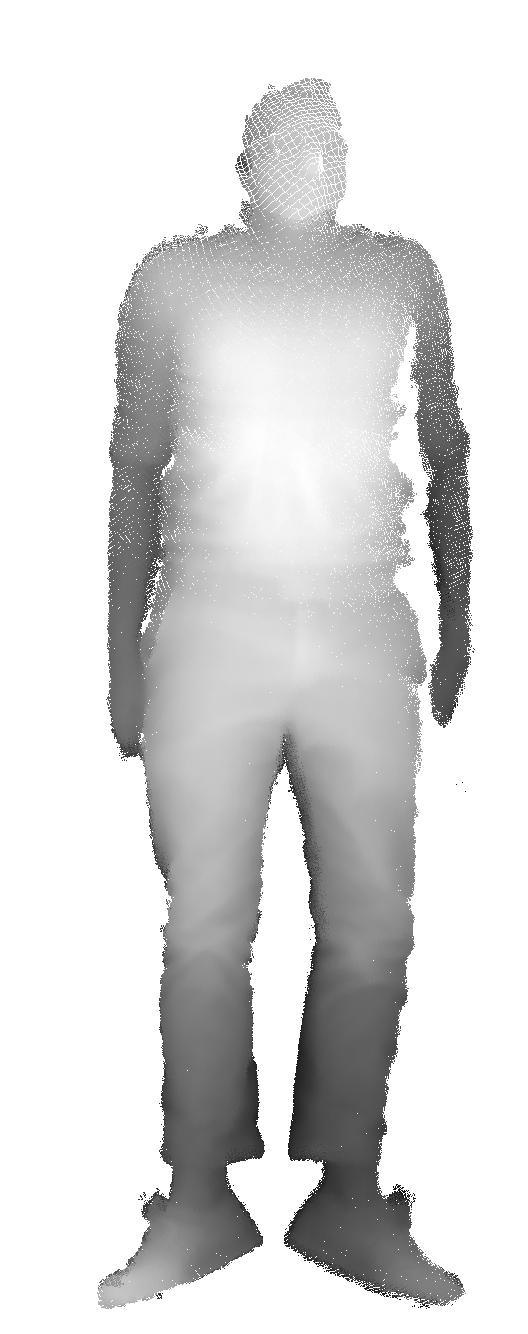}
        \includegraphics[width=1.8cm]{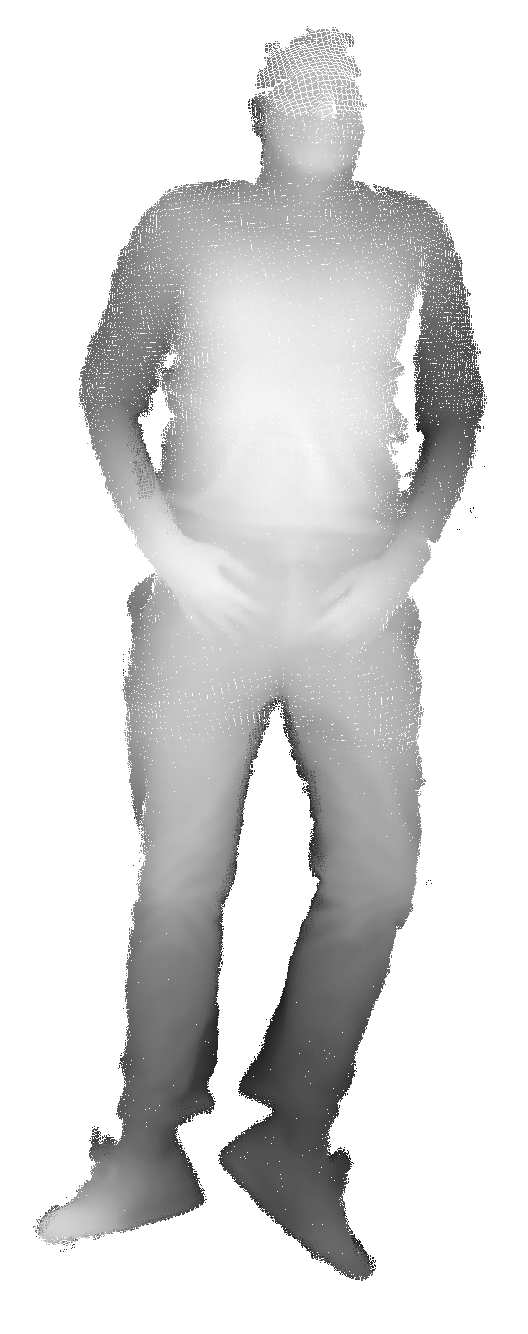}
        
        \caption{Heightmaps}
    \end{subfigure}
    \begin{subfigure}[t]{0.24\textwidth}
    \centering
        \includegraphics[width=1.8cm]{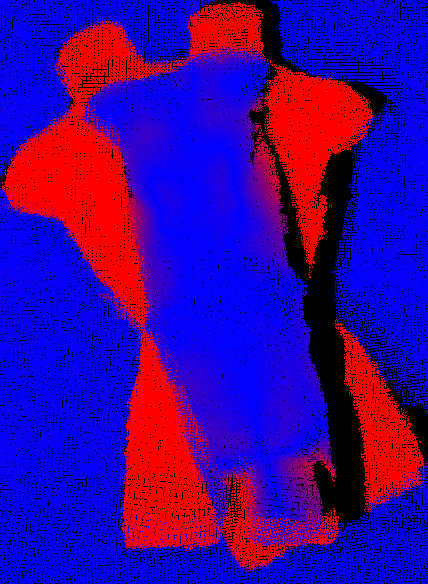}
        \includegraphics[width=1.8cm]{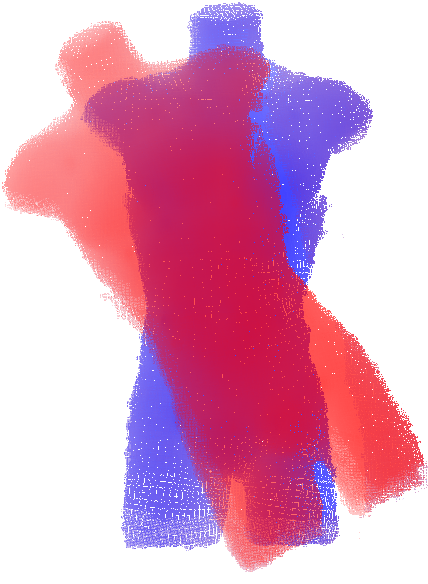}    
            
        \includegraphics[width=1.8cm]{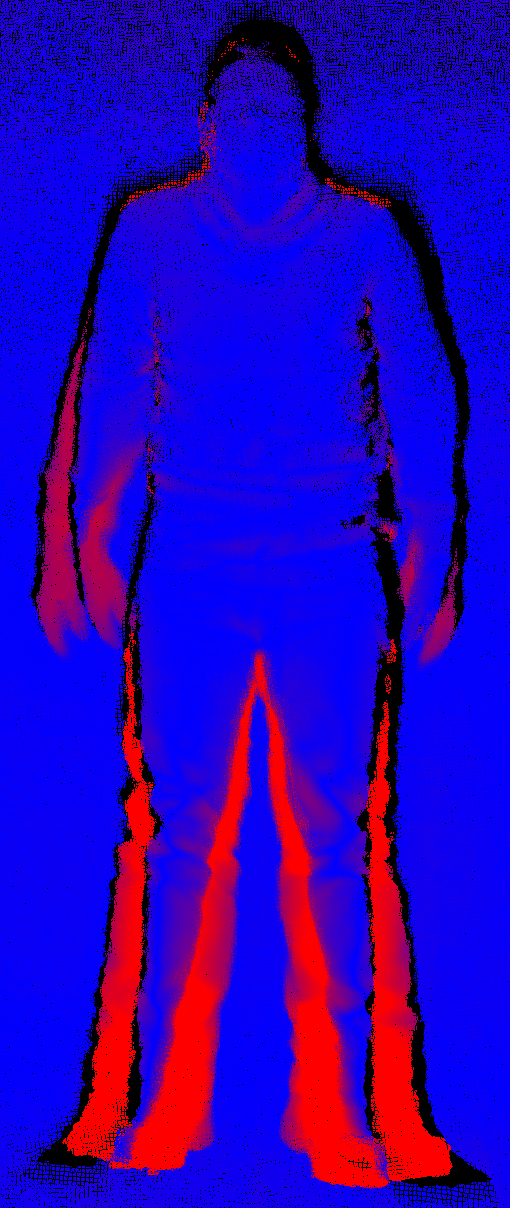}
        \includegraphics[width=1.8cm]{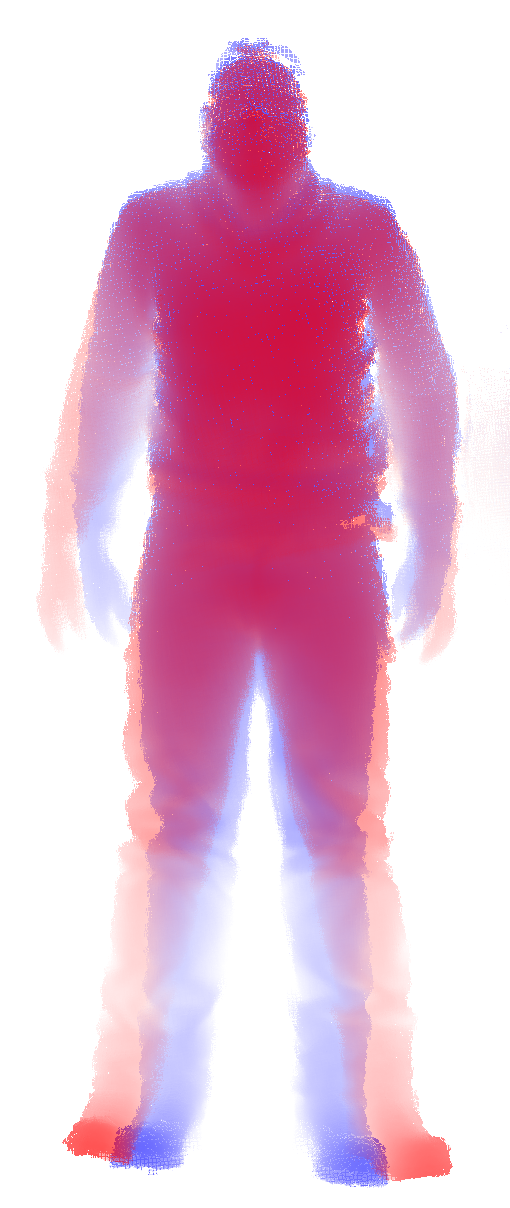}
        
        \includegraphics[width=1.8cm]{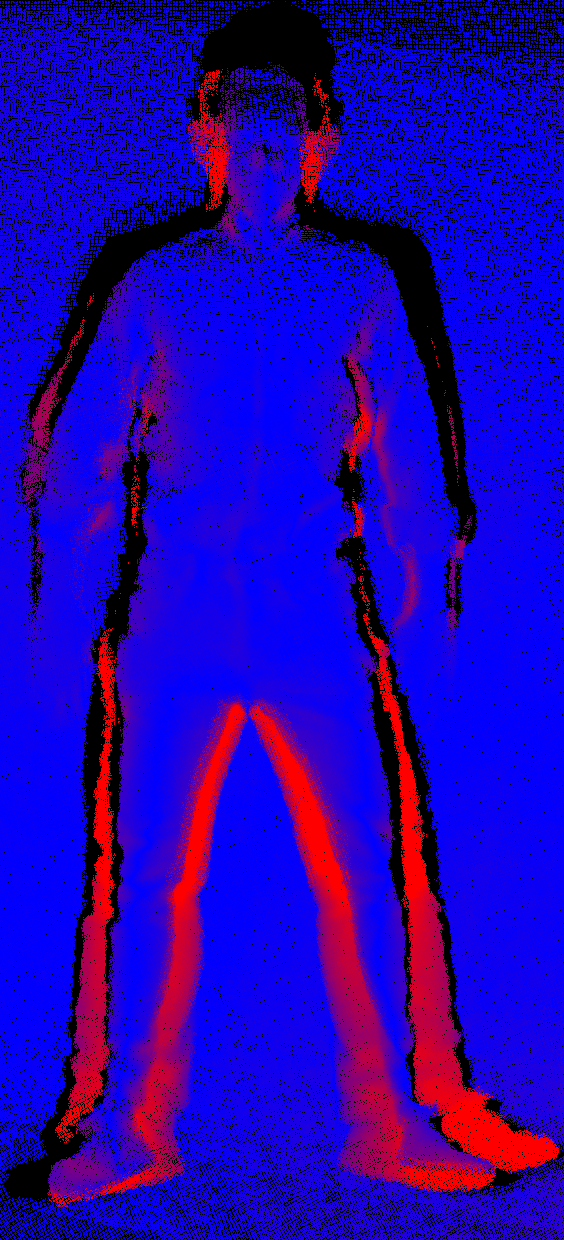}
        \includegraphics[width=1.8cm]{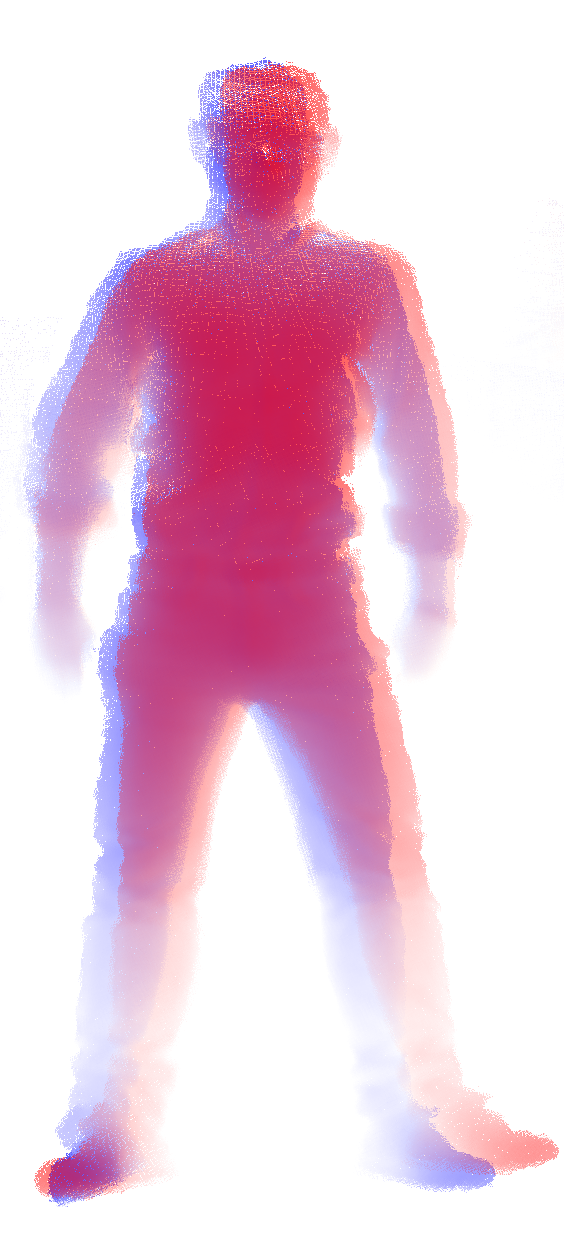}
        
        \includegraphics[width=1.8cm]{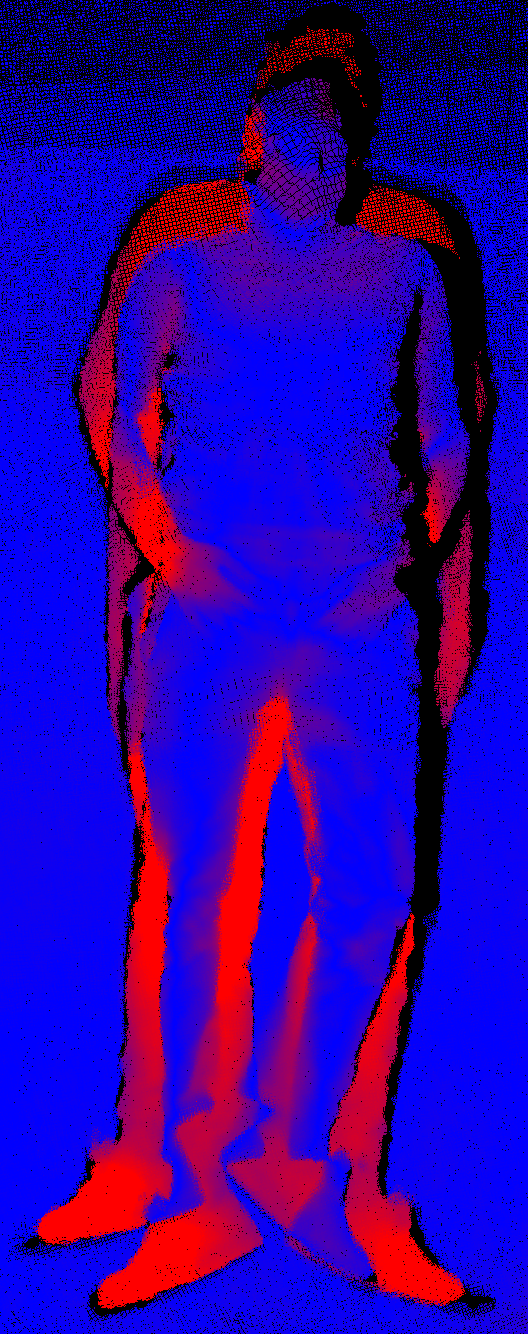}
        \includegraphics[width=1.8cm]{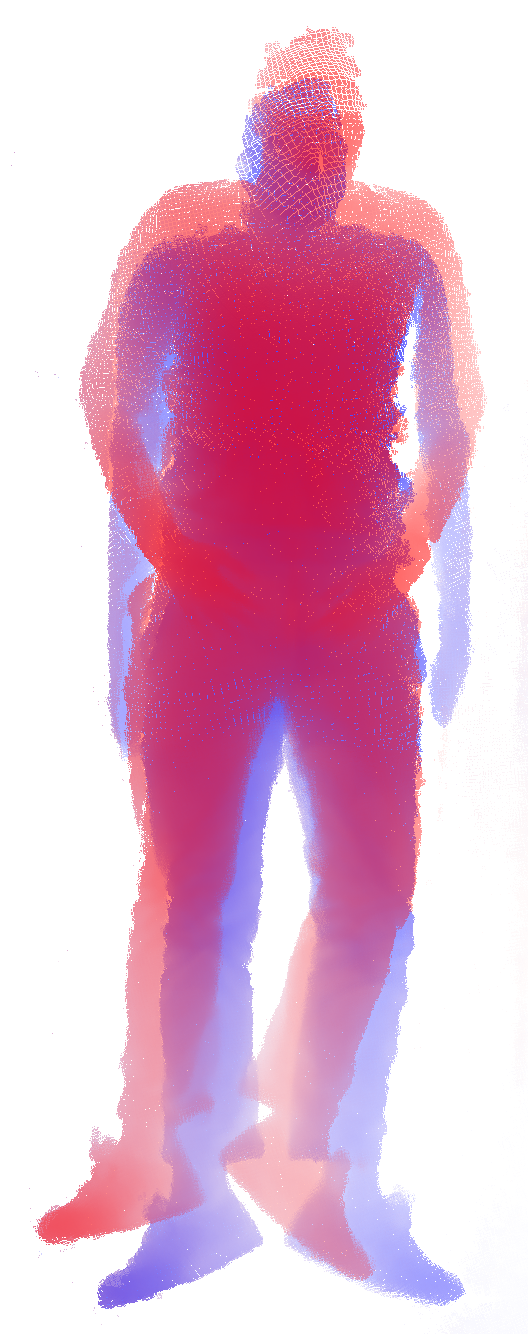}
        \caption{Error Heat Map and Overlay}
    \end{subfigure}
    \caption{Reconstructions made from our algorithm's output of the human subject in two difference scenes: a reference scene (a) and a new scene (b). The mesh reconstructions are done by applying the Poisson surface reconstruction algorithm on the point clouds made from the heightmap of the scene. Each mesh is presented twice first rendered with no color but with shading and second with color and no shading. The heightmaps from the two scenes can be seen in column (c). Finally, the error map of the second scene and overlay of the height maps are presented in column (d). Please note that in the heightmaps and their overlay, the patient is segmented for better visualization.}
    \label{fig:reconstructions}
\end{figure*}

Since the output of our system is a visual aid for patient positioning, this section  presents some qualitative results from the reconstructions obtained with our method. These results are made by calculating normals for each point in the heightmap, converting it to a point cloud, and finally applying the Poisson surface reconstruction algorithm done in the CloudCompare \footnote{cloudcompare.org} software. The images are obtained by rendering the mesh in the MeshLab \footnote{www.meshlab.net} software. These results can be seen in Figure \ref{fig:reconstructions}. 

The figure presents a different scenarios in each row, first the mannequin used in our quantitative evaluation and then  three different human subjects. In every row, first, the rendered reconstructed meshes for the reference scene and the new scene are displayed. For each scene the left image is rendered only using the mesh geometry and lighting, and the right image is the same mesh rendered with interpolated colors from the point cloud and no shading. After that, segmented heightmaps for the reference scene and the new scene could be observed. Finally, the error map and the image of overlayed heightmaps in different colors are presented. In the error map, blue shows an error of zero and red presents an error of \SI{10}{\centi\meter} or higher. The errors in between are shown by linearly interpolated colors between red and blue. In the image of overlayed heightmaps, the reference scene is colored in blue and the new scene is colored in red.

\begin{figure*}[t]
    \centering
    \includegraphics[width=\textwidth]{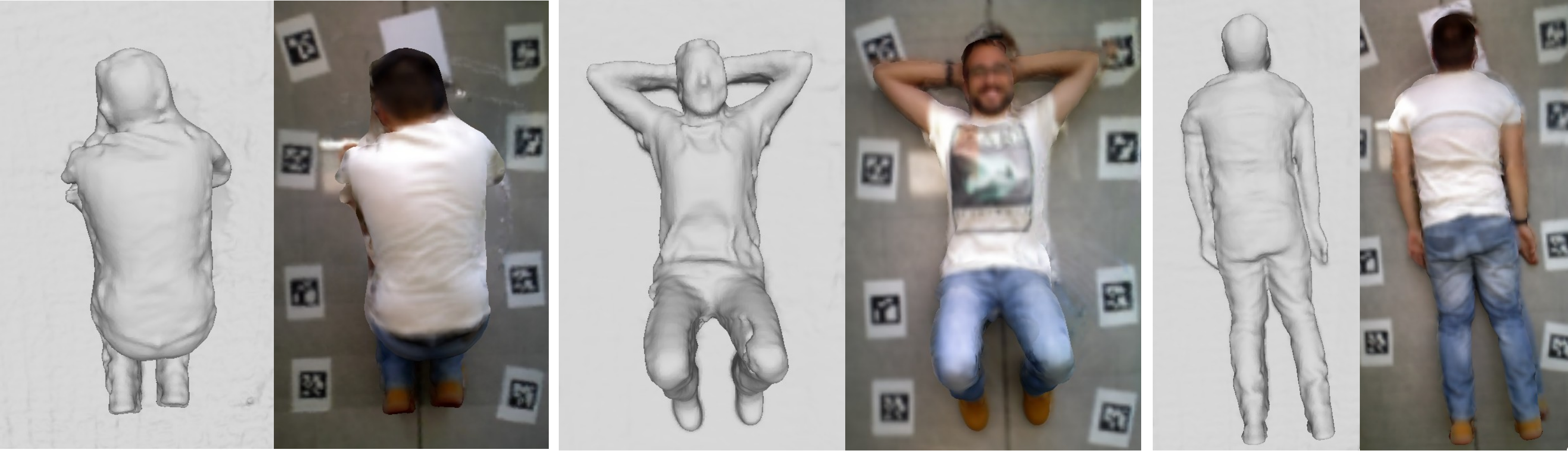}
    \caption{In the images above you can see reconstructions from the subject in different scenes taking different poses. Reconstructed meshes are created from the point clouds obtained from the output heightmaps of our algorithm. To do so, we have employed the Poisson surface reconstruction algorithm. You can see two renders of the mesh at each scene, using only lighting or using only shading. The other one using only the interpolated colors when reconstructing the mesh. Please note that the artifacts around the subject's body are due to lack of captured points in the point cloud and not the quality of the reconstruction.}
    \label{fig:reconst_poses}
\end{figure*}

More reconstructions could be seen in Figure \ref{fig:reconst_poses}. Here the subject takes different poses similar to those that are commonly used in radiation therapy. Sequences related to this sequence were captured with half the resolution of the ones in Figure \ref{fig:reconstructions}. 

\subsection{Computing time}
\label{subsec::computingtimes}
The proposed method is suitable for its integration in realistic environments using consumer grade equipment within a reasonable computing time, and without requiring dedicated graphic cards.

\begin{table}[]
    \centering
    \begin{tabular}{l|r}
         Algorithm Step & Time (\SI{}{\second}) \\
         \hline
         Extracting Pointclouds & 3.121\\
         Global ICP & 23.083\\
         Finding 3D Corner Positions & 0.003\\
         Scene Alignment* & 0.312\\
         Heightmap Creation & 3.903\\
         Error Map and Overlay Creation* & 0.256\\
         \hline
         Total & 30.679
    \end{tabular}
    \caption{Average computing time of our algorithm for different steps and in total.\\ *These steps are reported only including non-reference scenes because they do not apply to reference scenes in our algorithm.}
    \label{tab:computing_times}
\end{table}
Table \ref{tab:computing_times} shows the average computing times required by the different steps of our algorithm for the human sequences visualized in Figure \ref{fig:reconstructions}. The computation time were calculated on a laptop with with the Intel Core i7-4700HQ processor and $50$ iterations of global ICP. The Table does not include the computing time needed by the UcoSLAM algorithm for tracking because this method runs in real time while the RGB-D video is being recorded. On average the sequences had 15 keyframes produced by the UcoSLAM algorithm. The average length of recording for these sequences was only 10 seconds.

\section{Discussion}
\subsection{Quantitative evaluation}
As could be seen (Table \ref{tab:results_asus}), in total, the average positional error in estimating the pose of the mannequin is around \SI{11}{\milli\meter}. However, the median of the error is even lower at around \SI{10}{\milli\meter}. This suggests that for most of the scenes the positional error is lower than the average. The same could be seen as true for rotational error where a median of around \SI{1}{\degree} is achieved.
\subsection{Qualitative evaluation}
As can be observed in Figure \ref{fig:reconstructions}, the human subjects are reconstructed with a considerable details. The same can be said for the shape of the mannequin; even the small IR reflective markers attached on the mannequin are clearly reconstructed. Please note that the artifacts on the edge of the person and the mannequin are due to the lack of captured points in those areas and not low quality of the point clouds. Furthermore, the last two images (in the far right) for each subject can clearly display the amount of error and how the patient needs to be moved to correct the pose. This is a desired feature since these images are shown to the person in charge of patient positioning as a guide for pose correction.

However as can bee seen in Figure \ref{fig:reconst_poses} the reconstructions are still robust. Again, please note that the artifacts on the edge of the subject are due to lack of the captured 3D points and not the quality of the reconstruction.
\subsection{Computing time}
As can be observed (Table \ref{tab:computing_times}), the most time consuming part of our implementation is the Global ICP. Nevertheless, in less than half a minute, our method is able to produce its results. We find this suitable for its use in real radiotherapy sessions.
\section{Conclusion}

This paper has proposed a novel approach to obtain 3D body reconstructions for patient positioning using inexpensive consumer RGB-D cameras. The main novelty of our approach is the use of a novel SLAM technique that combines natural and artificial landmarks in order to obtain a coarse 3D reconstruction that is later improved without requiring expensive hardware or dedicated graphic cards.

By placing a set of squared markers in the environment, that should remain fixed from one session to another, the proposed method is able to align the reconstructions achieving a median translation error of \SI{1}{\centi\meter} and a rotational error of \SI{1}{\degree}. 

The use of markers also allows us to employ the RGB-D camera as a hand-held scanner. Thus, the recording distance to the patient is reduced contributing to improve the reconstruction quality for that type of sensors. Our method generates as output a visual superimposition of the patient both in its current position, and in the reference position, along with an error map. These pieces of information allow us to easily check how the pose of the patient needs to be corrected.

We have created a robust framework that can produce results of decent quality and could be improved by enhancing different parts of it. We suggest that this method could be extended for non-rigid surface registration which is left to be done in future works. 

Our method in this paper is focused on using a single CPU so that it is usable in most of the situations. However with the help of GPU accelerated computing it is possible to speed up our method and also include non-rigid registration with a reasonable computation time. Furthermore it is also possible to use multiple consumer level RGB-D sensor on a camera rig to improve the quality of the scan. This can give the opportunity of fixing the camera rig in the room and have a live view of the patient which could be valuable in monitoring breath and live registration of the CT scan on the patient's body.

\section*{Conflict of interest}

The authors do not have financial and personal relationships with other people or organizations that could inappropriately influence (bias) their work.

\section*{Acknowledgment}
  This project has been funded under projects TIN2016-75279-P and IFI16/00033 (ISCIII) of Spain Ministry of Economy, Industry and Competitiveness, and FEDER. The authors thank the Health Time Radiology Company for its help for the Project IFI16/00033 (ISCIII)

\bibliographystyle{ieeetr}
\bibliography{references}
\newpage
\appendix

\end{document}